\pdfoutput=1

\documentclass[11pt]{article}

\usepackage[]{acl}

\usepackage{times}
\usepackage{latexsym}

\usepackage[T1]{fontenc}

\usepackage[utf8]{inputenc}

\usepackage{microtype}

\usepackage{inconsolata}

\usepackage{graphicx}
\usepackage{booktabs}
\usepackage{array} 
\usepackage{times}
\usepackage{latexsym}
\usepackage{paralist}
\usepackage{enumitem}
\usepackage{comment}
\newcommand{\hide}[1]{}

\title{ChronoSense: Exploring Temporal Understanding in Large Language Models with Time Intervals of Events}

\author{Duygu Sezen Islakoglu$^*$ \\
  Utrecht University, Netherlands \\
  \texttt{d.s.islakoglu@uu.nl} \\\And
  Jan-Christoph Kalo$^*$\\
  University of Amsterdam, Netherlands\\
  \texttt{j.c.kalo@uva.nl} \\
}

\begin{document}
\maketitle
\def\thefootnote{*}\footnotetext{Equal contribution} \def\thefootnote{\arabic{footnote}}

\begin{abstract}
Large Language Models (LLMs) still face significant challenges in reasoning and arithmetic. Although temporal reasoning has raised increasing research attention, comprehensive testing of Allen's interval relations (e.g., before, after, during) --a fundamental framework for temporal relationships-- remains underexplored.
To fill this gap, we present \textit{ChronoSense}, a new benchmark for evaluating LLMs' temporal understanding. 
It includes 16 tasks, identifying the Allen relation between two temporal events and temporal arithmetic.
We assess the performance of seven recent LLMs.
The results indicate that models handle Allen relations, even symmetrical ones, quite differently. 
Moreover, the findings suggest that the models may rely on memorization to answer time-related questions. Overall, the models' low performance highlights the need for improved temporal understanding in LLMs. 
Our dataset and the source code are available at 
\url{https://github.com/duyguislakoglu/chronosense}.

\end{abstract}
\section{Introduction}
Large Language Models (LLMs) have demonstrated remarkable proficiency across various tasks in NLP. 
Despite these advancements, significant challenges persist in areas such as reasoning, arithmetic~\cite{srivastava2023beyond}, and working with numerical values~\cite{wei2022chain}. 
These limitations affect their performance in temporal reasoning and numerical arithmetic.

Recent research has shown a growing interest in evaluating the temporal reasoning capabilities of LLMs. Efforts have focused on event ordering, comparing temporal events, temporal question answering, and event forecasting~\cite{Chu2023TimeBenchAC}. However, a notable gap remains: the comprehensive testing of Allen's intervals, one of the most fundamental temporal reasoning frameworks that have been in use for over 30 years~\cite{10.5555/93913.93994}.

\begin{figure}
    \centering
\captionsetup{justification=centering,margin=1cm}
    \includegraphics[width=0.37\textwidth]{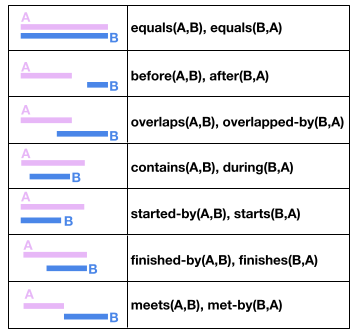}
        \caption{13 Allen relations between two intervals, covering all combinations.}
    \label{allen} 
\end{figure}

Allen's intervals provide a formal structure for representing temporal relationships between events, defining thirteen possible relations between time intervals. Despite its importance, existing benchmarks cover only subsets of these relations. We demonstrate these 13 relations in Figure~\ref{allen}.

\begin{figure}
    \centering
\captionsetup{justification=centering,margin=1cm}
    \includegraphics[width=0.3\textwidth]{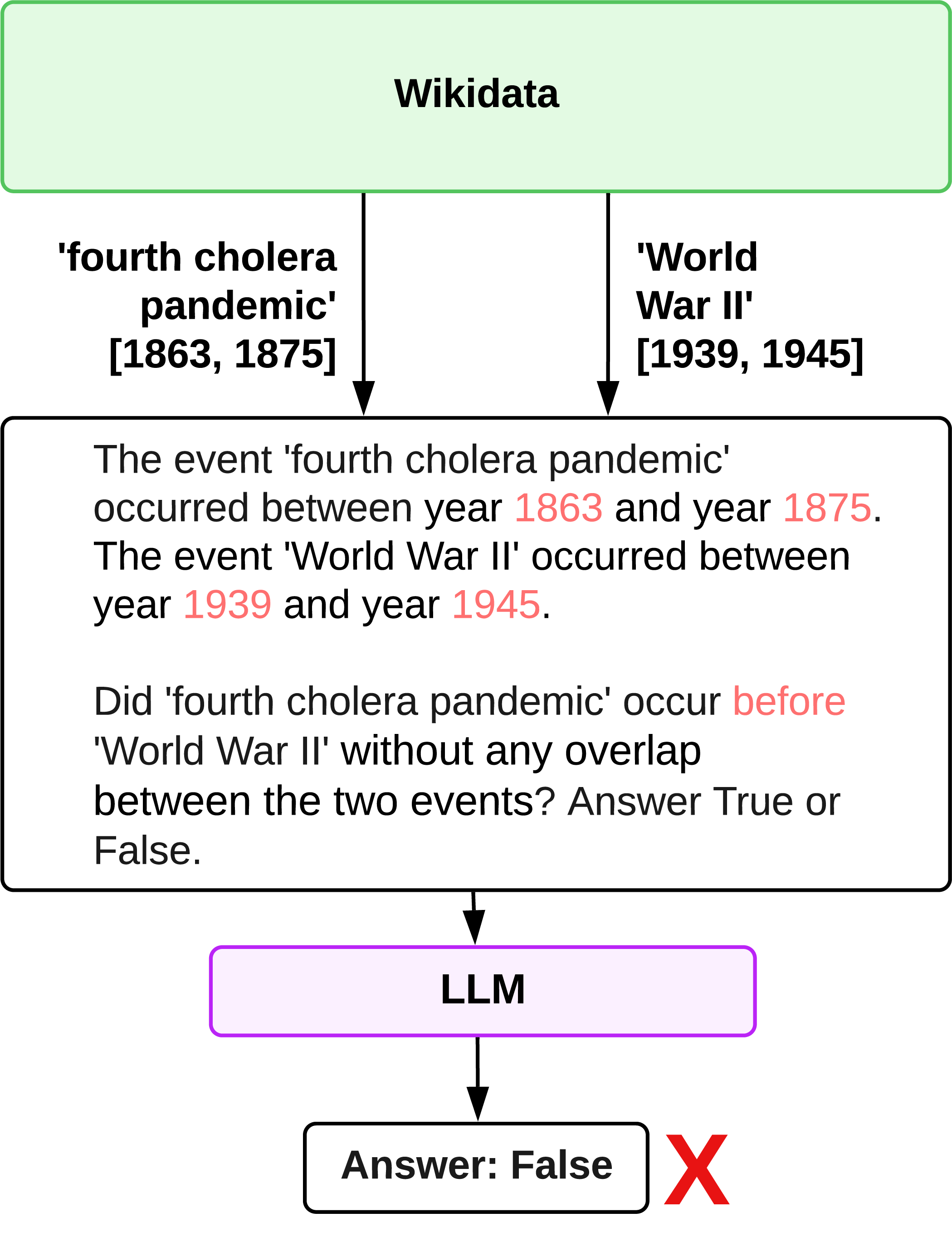}
    \caption{An example for comparing two temporal events with LLMs.}
    \label{overview} 
\end{figure}

To illustrate our task, consider the following example:
In Figure~\ref{overview}, the first event is the \textit{fourth cholera pandemic} which occurred between \texttt{1863} and \texttt{1875}, while \textit{World War II} occurred between \texttt{1939} and \texttt{1945}. In our prompt, we list these two events with their names and respective start and end years and then ask a \textit{True/False} question about one of the 13 Allen relations. For example, we ask the LLM whether the \textit{fourth cholera pandemic} happened "before" \textit{World War II}.

While such tasks are straightforward for humans, they pose considerable difficulty for LLMs due to the need to compare numerical values accurately.
Our research focuses on reasoning about time intervals and assessing how models perform on temporal understanding tasks. We also incorporate three time arithmetic tasks to challenge the models further. 

Our contributions can be summarized as follows:
\begin{itemize}[noitemsep]
    \item We present a comprehensive evaluation of LLMs' performance on temporal reasoning tasks using our \textit{ChronoSense} benchmark. Our evaluation spans Allen relations and temporal arithmetic tasks across 0-shot, few-shot, and chain-of-thought (CoT) prompting scenarios.
   \item We demonstrate the effectiveness of few-shot and CoT prompting in improving LLM performance, especially on temporal arithmetic tasks that require step-by-step reasoning.
    \item We investigate the influence of memorization on LLMs' ability to perform temporal reasoning tasks, especially when models encounter real-world event names that might have been part of pre-training data.
\end{itemize}

\section{Preliminaries}
\textbf{Allen's Interval Algebra}.
Allen's interval algebra (IA) \citep{10.5555/93913.93994}
provides 13 different relations between two intervals. 
As illustrated in Figure \ref{allen}, these relations are "Equals", "Before", "After", "Overlaps", "Overlapped-by", "Contains", "During",
"Started-by", "Starts", "Finished-by", "Finishes" 
"Meets" and "Met-by". 
These relations are mutually exclusive and cover all possible temporal relationships between two intervals. IA serves as a base for artificial intelligence and has been used in many applications \citep{janhunen2019allensintervalalgebramakes}. Although it is not the focus of this study, it allows deriving new facts. For instance, through transitivity, if Event $e_1$ happens before Event $e_2$, and Event $e_2$ happens before Event $e_3$, then Event $e_1$ happens before Event $e_3$. Therefore, correctly identifying the relationships between intervals is essential to support this type of reasoning.
\section{ChronoSense Dataset}
\label{methodology}
We create an event-centric dataset, named \textit{ChronoSense}\footnote{The dataset will be released under the CC BY 4.0 license.}. This dataset is designed to diagnose how well LLMs comprehend temporal events and the relationships between them, as illustrated in Figure~\ref{overview}. ChronoSense contains True/False questions that include different temporal dimensions. It features two types of questions: 
\begin{inparaenum}[(1)]
 \item Allen questions (requiring models to determine the Allen relation of two time intervals) and 
 \item temporal arithmetic tasks focused on a single event (challenging models to draw conclusions based on explicit time information).   
\end{inparaenum}
We set the time granularity to years for both question types. 
The prompts used in ChronoSense can be seen in Table \ref{templates} in Appendix \ref{Appx}.

\noindent
\textbf{Question Type 1: Comparing Two Temporal Events with Allen Relations.}
We extract real event pairs from the Wikidata \citep{vrandevcic2014wikidata} (Section \ref{allen-generation}). Similar to \citep{yang2023logical}, every test instance in our dataset is in (Context, Hypothesis, Correctness) format. \textit{Context} introduces the events and explicitly states the time periods when the events have occurred  (e.g. \textit{The event ‘fourth cholera pandemic’ occurred between year 1863 and year 1875. The event ‘World War II’ occurred between year 1939 and year 1945.}). \textit{Hypothesis} verbalizes an Allen relation in natural language (e.g. \textit{Did ‘fourth cholera pandemic’ occur before ‘World War II’ without any overlap between the two events? Answer True or False.}). \textit{Correctness} is True if \textit{Hypothesis} describes the temporal relationship between these two events correctly and False otherwise (e.g. \textit{True} for the example above.). 

\noindent
\textbf{Question Type 2: Temporal Arithmetic With A Single Event.}
To get insights into models' ability to perform temporal arithmetic, we also include temporal arithmetic questions in \textit{ChronoSense}.
\textit{Context} introduces a single event and explicitly states the time information and a temporal feature such as its duration or frequency (e.g. \textit{‘Event A' first occurred in year 1909. ‘Event A' occurs every 12 years.}). 
\textit{Hypothesis} is a statement that is not covered in \textit{Context} and requires arithmetic calculations to verify  (e.g. \textit{Did ‘Event A' occur again in the year 1921? Answer True or False.}).
\textit{Correctness} is True if \textit{Hypothesis} matches with the calculations based on the \textit{Context} and False otherwise (e.g. \textit{True} for the example above). 

The temporal arithmetic questions cover three different aspects. \textit{End Timepoint} focuses on the duration of an event and requires models to determine the end time based on the given start time and duration.
\textit{Next Occurrence} focuses on the frequency of events and challenges models to calculate when an event occurs again based on a given frequency.
\textit{Intermediate Timepoint}, which is novel to this work, challenges models to infer whether an event was happening between its start and end time by asking if it happened at a certain year in time. Due to the limited number of events with frequency from Wikidata, we synthetically create these questions. 
Therefore, the events do not have event names, but rather we name them as "Event A". For each question, we create a negative sample by creating a wrong \textit{Hypothesis} (e.g. by changing the next occurrence year in the previous example from 1921 to 1950.).\\
\textbf{Different event abstraction levels.}
For Allen questions, we have an abstract version of each question where we hide the names of the events by replacing them with letters such as "Event A" and "Event B". This setting allows us to see how the memorization affects LLM's performance by comparing the abstract versions with the original versions (where we have event names).\\
\textbf{Different prompts for questions.}
There are multiple ways to ask a question, so we create two different additional prompts for each question to understand the effect of the prompt. All prompts can be seen in Table \ref{templates} and Table \ref{prompt-type-templates} in the Section \ref{Appx}.\\
\textbf{Negative samples.}
To evaluate the robustness of the LLM's predictions, we generate negative examples for each data instance (detailed in \ref{negative-sample}). Therefore, the  \textit{Correctness} value is "\textit{True}" in 50\% of the data instances, and "\textit{False}" in the other half. 

\noindent
\textbf{Dataset statistics.}
For each Allen relation and each temporal arithmetic question, ChronoSense has 4,000 training samples, 500 validation samples, and 500 test samples.
\section{Experiments}  
We evaluate the performance of various LLMs on a task framed as binary classification. Specifically, the models are tasked with answering \textit{True} or \textit{False} to a set of prompts on temporal reasoning. We evaluate the accuracy of the models, where we have a random chance accuracy of 50\%. We compare the following LLMs in our experiments:
Gemma2-9b-it\footnote{\url{https://huggingface.co/google/gemma-2-9b-it}},
GPT-4o {(gpt-4o-2024-05-13)\footnote{\url{https://platform.openai.com/docs/models/gpt-4o}},
GPT-4o-mini (gpt-4o-mini-2024-07-18)\footnote{\url{https://platform.openai.com/docs/models/gpt-4o-mini}},
Meta-Llama-3.1-8B-Instruct\footnote{\url{https://huggingface.co/meta-llama/Meta-Llama-3.1-8B-Instruct}},
Mistral-7B-Instruct-v0.2\footnote{\url{https://huggingface.co/mistralai/Mistral-7B-Instruct-v0.2}} \cite{jiang2023mistral7b},
Mixtral-8x7B-Instruct-v0.1\footnote{\url{https://huggingface.co/mistralai/Mixtral-8x7B-Instruct-v0.1}} \cite{jiang2024mixtralexperts},
Phi-3-mini-128k-instruct\footnote{\url{https://huggingface.co/microsoft/Phi-3-mini-128k-instruct}}. 
Each model can generate up to 64 new tokens for an answer; however, in the chain-of-thought (CoT) setting, the maximum token limit is increased to 512 to provide more space for reasoning. For both question types (Allen and temporal arithmetic), we report on different settings: 0-shot, 1-shot, 3-shot, and Chain-of-Thought (CoT) prompting. For CoT experiments, we add a "Let's think step by step." sentence to the original prompts (Table \ref{templates}). This follows the idea introduced in \citep{NEURIPS2022_8bb0d291}. For Allen questions, we also report on abstract versions in which we remove the real event names. As mentioned in Section \ref{methodology}, the temporal arithmetic questions are all in the abstract setting. 

We report the averaged results in Table \ref{main-table}. The complete experimental results, including the experiments on individual Allen relations, can be found in \ref{appx:detailed-results}. Moreover, in Table \ref{0-shot-gpt4o-mixtral-phi}, we zoom in and report the 0-shot performance on individual Allen relations for three models. 
We include qualitative examples of failure cases in Section~\ref{failure-cases}, and provide an analysis of different prompt variants in Section~\ref{app-diff-prompts}.

\begin{table*}[t]
\resizebox{\textwidth}{!}{%
\begin{tabular}{llccccccrr}
\hline
\multicolumn{1}{l}{Type} & \multicolumn{1}{l}{Setting}  &\multicolumn{1}{c}{Gemma2-9B-it} 
&\multicolumn{1}{c}{GPT-4o} &\multicolumn{1}{c}{GPT-4o-mini} & \multicolumn{1}{c}{Llama3.1-8B} & \multicolumn{1}{c}{Mistral-7B} & \multicolumn{1}{c}{Mixtral-8x7B} & \multicolumn{1}{r}{Phi-3-mini} \\
\hline
Allen & 0-shot & 0.09* & 0.87 & 0.72 & 0.13* & 0.50 & 0.54 & 0.56 \\
&1-shot & 0.75 & 0.93 & 0.75 & 0.01* & 0.47 & 0.56 & 0.59 \\ 
&3-shot & 0.26* & 0.95 & 0.78 & 0.01* & 0.49 & 0.58 & 0.66\\ 
&CoT & 0.75 & 0.65 & 0.69 & 0.75 & 0.51 & 0.57 & 0.75 \\ 
&Abstract & 0.15* & 0.78 & 0.64 & 0.14* & 0.23* & 0.35 & 0.61  & \\
\hline
Arithmetic&0-shot &  0.76 & 0.55 & 0.60 & 0.48 & 0.36* & 0.35 & 0.67\\
&1-shot & 0.71 & 0.50* & 0.49 & 0.19 & 0.43 & 0.50 & 0.65 \\
&3-shot &0.54 & 0.51 & 0.39* & 0.10* & 0.47 & 0.64 & 0.37 \\
&CoT&  0.94 & 0.99 & 0.99 & 0.92 & 0.70 & 0.75 & 0.98\\
\hline
\end{tabular}
}
 \caption{The average performance comparison between different settings on two different question types in ChronoSense. (*) indicates the models that perform poorly due to producing a high number of unclear answers ($\geq$ 250) in the majority of the tasks. These models fail to follow the instruction by not answering with "True" or "False" as required.}
\label{main-table}
\end{table*}

\begin{table}[t]
\resizebox{.48\textwidth}{!}{%
\begin{tabular}{lccr}
\hline
\multicolumn{1}{l}{Allen Relation} &\multicolumn{1}{c}{GPT-4o} &\multicolumn{1}{c}{Mixtral-8x7B} & \multicolumn{1}{c}{Phi-3-mini} \\
\hline
After & 0.956 & 0.78 & 0.566 \\
Before & 0.914 & 0.902 & 0.758 \\
Contains & 0.884 & 0.472 & 0.652 \\
During & 0.878 & 0.512 & 0.49 \\
Equals & 0.69 & 0.336 & 0.54 \\
Finished-By & 0.926 & 0.398 & 0.486 \\
Finishes & 0.908 & 0.43 & 0.492 \\
Meets & 0.91 & 0.74 & 0.488 \\
Met-By & 0.864 & 0.594 & 0.494 \\
Overlapped-By & 0.842 & 0.476 & 0.786 \\
Overlaps & 0.884 & 0.43 & 0.648 \\
Started-By & 0.896 & 0.578 & 0.474 \\
Starts & 0.846 & 0.442 & 0.492 \\
\hline
\end{tabular}
}
\caption{0-shot setting results for GPT-4o, Mixtral-8x7B, and Phi-3-mini on 13 Allen relations.}
\label{0-shot-gpt4o-mixtral-phi}
\end{table}

\noindent
\textbf{General Findings.}
\begin{inparaenum}[(1)] 
    \item The models exhibit low performance and lack consistency on ChronoSense questions across the experiments, given the fact that the random prediction would lead to 0.50 accuracy.  This suggests the need for improvements in temporal understanding in LLMs.
    \item Few-shot and CoT settings are helpful for most models for Allen questions. Despite these improvements, the tasks remain challenging, as several models still have an accuracy below 0.60.
    \item Arithmetic questions are typically more challenging than Allen relations in both zero-shot and few-shot settings.
    For these questions, the few-shot setting only improves Mistral-7B and Mixtral-8x7B models. However, CoT prompting enhances model performance on arithmetic questions across all models. This is expected as these questions require step-by-step reasoning.
    \item When averaged over models, some Allen relations are easier and some are more challenging for the models. First, "Before" and "After" are easier than other relations in all experiments, with one exception. This is expected as these relations are the most frequently used phrases among others. This may also indicate that the models are better at detecting relations that do not contain any overlap. Second, "Equals" is the hardest relation in zero-shot and CoT settings, and "Finishes" is the hardest for few-shot and abstract settings. The questions for both relations require checking whether the endpoints of events are the same. 
    \item The models do not perform similarly for symmetrical Allen relations. For instance, despite their symmetric nature, the averaged model performance for "Before" is higher than for "After" and "Meets" is higher than "Met-by".
    Similarly, "Contains", "Finished-by" and "Overlaps" are easier than their symmetrical relations ("During", "Finishes" and "Overlapped-by") with one exception.
   \item The abstract versions are more challenging for most of the models. Models may rely on memorization to answer temporal understanding questions for the events included in the pre-training data. In other words, the implicit knowledge from pre-training can influence their performance on temporal understanding.
   \item As illustrated in Section \ref{failure-cases}, the types of model failures include: confusion between start and end years, incorrect reasoning, calculation errors (including extra calculations), incorrect conclusions despite correct explanations, and confusion caused by temporal granularity.
\end{inparaenum}

\section{Related Work}
Temporal reasoning has been extensively studied in NLP~\cite{terenziani2009,sanampudi2010temporal} and  QA over temporal knowledge graphs~\cite{dhingra-etal-2022-time,zhao-etal-2022-language,saxena-etal-2021-question,chen2021dataset,jia2018,jia2018-tequila,jia2021complex}.
A new line of work focuses on LLMs' temporal knowledge and reasoning. TimeBench~\cite{Chu2023TimeBenchAC} covers abstract temporal expressions, commonsense reasoning, and event relationships. Other benchmarks include those by~\cite{jain2023language} for commonsense-based temporal tasks and TimeLlama~\cite{yuan2023future} for event forecasting. 
TGQA~\cite{xiong2024large} evaluates synthetic temporal QA but only covers three simple event relations. TRACIE~\cite{zhou-etal-2021-temporal} assesses reasoning over implicit events, while TEMPREASON~\cite{Tan2023TowardsBA} probes three levels of temporal understanding but primarily focuses on factual recall. TRAM~\cite{Wang2023TRAMBT} includes event relations from~\cite{uzzaman-etal-2013-semeval} but lacks explicit events. ~\cite{tan-etal-2023-towards} has temporal arithmetic but it is event-independent. LTLBench~\cite{Tang2024LTLBenchTB} uses linear temporal logic to model the temporal relationships between events. Test of Time~\cite{fatemi2024testtimebenchmarkevaluating} creates a synthetic dataset to isolate temporal reasoning. Recent works on event ordering include TDDiscourse~\cite{naik-etal-2019-tddiscourse}, which classifies implicit event relations overlapping with Allen's framework. Datasets from~\cite{vashishtha-etal-2020-temporal} focus on event ordering and duration, while TORQUE~\cite{ning-etal-2020-torque} presents a reading comprehension dataset to investigate the temporal ordering
of events but lacks explicit start and end times. Despite the variety of benchmarks, none covers all 13 of Allen’s interval relations.

\section{Conclusion}   
We introduce ChronoSense, a diagnostic dataset designed to assess LLMs' ability to compare event timelines using Allen relations and perform temporal arithmetic. We show that models frequently struggle with these tasks and may rely on memorization rather than reasoning. This raises critical concerns about their reliability in applications such as historical analysis, legal AI, and medical timelines. Future research should focus on improving LLMs' temporal reasoning capabilities, integrating temporal constraint-based reasoning, and analyzing multi-event comparisons.

\section{Limitations}
Our work has some limitations regarding the dataset and the evaluation. Concerning the dataset, some Wikidata events have ambiguous names that may mislead the model, e.g., an exhibition event named after a painter, which may not clearly indicate a temporal event to the model. 
On the evaluation side, our study involves a relatively small selection of models and some closed-source models (e.g. GPT-4o). Moreover, although we test 3 different prompt versions per task, we acknowledge that the prompt content may influence the model's performance. Lastly, we truncate the LLM outputs when they exceed the maximum token lengths. This potentially omits some of the correct answers and leads to lower accuracy scores for the respective models.

\section{Ethics Statement}
Our dataset, which sources events from Wikidata, inherently carries the risk of containing incorrect information. This could unintentionally propagate misinformation. While our script filters out data points containing certain triggering keywords, some event names may still include inappropriate or harmful content. This does not reflect the views or opinions of the authors. Moreover, the data points in ChronoSense do not represent individuals but rather events categorized as instances or subclasses of "occurrence" \footnote{\url{https://www.wikidata.org/wiki/Q1190554}}. However, some events include the names of individuals, such as exhibitions named after artists. Furthermore, we acknowledge the environmental impact associated with LLMs. Although our study only utilizes pre-trained models, inference with these models still demands significant computational resources.
\bibliography{references}

\begin{thebibliography}{32}
\providecommand{\natexlab}[1]{#1}

\bibitem[{Allen(1989)}]{10.5555/93913.93994}
James~F. Allen. 1989.
\newblock \emph{Maintaining Knowledge about Temporal Intervals}, page 361–372.

\bibitem[{{BIG-bench authors}(2023)}]{srivastava2023beyond}
{BIG-bench authors}. 2023.
\newblock \href {https://openreview.net/forum?id=uyTL5Bvosj} {Beyond the imitation game: Quantifying and extrapolating the capabilities of language models}.
\newblock \emph{Transactions on Machine Learning Research}.

\bibitem[{Chen et~al.(2021)Chen, Wang, and Wang}]{chen2021dataset}
Wenhu Chen, Xinyi Wang, and William~Yang Wang. 2021.
\newblock \href {https://arxiv.org/abs/2108.06314} {A dataset for answering time-sensitive questions}.
\newblock \emph{Preprint}, arXiv:2108.06314.

\bibitem[{Chu et~al.(2023)Chu, Chen, Chen, Yu, Wang, Liu, and Qin}]{Chu2023TimeBenchAC}
Zheng Chu, Jingchang Chen, Qianglong Chen, Weijiang Yu, Haotian Wang, Ming Liu, and Bing Qin. 2023.
\newblock Timebench: A comprehensive evaluation of temporal reasoning abilities in large language models.
\newblock \emph{ArXiv}, abs/2311.17667.

\bibitem[{Dhingra et~al.(2022)Dhingra, Cole, Eisenschlos, Gillick, Eisenstein, and Cohen}]{dhingra-etal-2022-time}
Bhuwan Dhingra, Jeremy~R. Cole, Julian~Martin Eisenschlos, Daniel Gillick, Jacob Eisenstein, and William~W. Cohen. 2022.
\newblock \href {https://doi.org/10.1162/tacl_a_00459} {Time-aware language models as temporal knowledge bases}.
\newblock \emph{Transactions of the Association for Computational Linguistics}, 10:257--273.

\bibitem[{Fatemi et~al.(2024)Fatemi, Kazemi, Tsitsulin, Malkan, Yim, Palowitch, Seo, Halcrow, and Perozzi}]{fatemi2024testtimebenchmarkevaluating}
Bahare Fatemi, Mehran Kazemi, Anton Tsitsulin, Karishma Malkan, Jinyeong Yim, John Palowitch, Sungyong Seo, Jonathan Halcrow, and Bryan Perozzi. 2024.
\newblock \href {https://arxiv.org/abs/2406.09170} {Test of time: A benchmark for evaluating llms on temporal reasoning}.
\newblock \emph{Preprint}, arXiv:2406.09170.

\bibitem[{Jain et~al.(2023)Jain, Sojitra, Acharya, Saha, Jatowt, and Dandapat}]{jain2023language}
Raghav Jain, Daivik Sojitra, Arkadeep Acharya, Sriparna Saha, Adam Jatowt, and Sandipan Dandapat. 2023.
\newblock Do language models have a common sense regarding time? revisiting temporal commonsense reasoning in the era of large language models.
\newblock In \emph{Proceedings of the 2023 Conference on Empirical Methods in Natural Language Processing}, pages 6750--6774.

\bibitem[{Janhunen and Sioutis(2019)}]{janhunen2019allensintervalalgebramakes}
Tomi Janhunen and Michael Sioutis. 2019.
\newblock \href {https://arxiv.org/abs/1909.01128} {Allen's interval algebra makes the difference}.
\newblock \emph{Preprint}, arXiv:1909.01128.

\bibitem[{Jia et~al.(2018{\natexlab{a}})Jia, Abujabal, Saha~Roy, Str\"{o}tgen, and Weikum}]{jia2018}
Zhen Jia, Abdalghani Abujabal, Rishiraj Saha~Roy, Jannik Str\"{o}tgen, and Gerhard Weikum. 2018{\natexlab{a}}.
\newblock \href {https://doi.org/10.1145/3184558.3191536} {Tempquestions: A benchmark for temporal question answering}.
\newblock In \emph{Companion Proceedings of the The Web Conference 2018}, WWW '18, page 1057–1062.

\bibitem[{Jia et~al.(2018{\natexlab{b}})Jia, Abujabal, Saha~Roy, Str\"{o}tgen, and Weikum}]{jia2018-tequila}
Zhen Jia, Abdalghani Abujabal, Rishiraj Saha~Roy, Jannik Str\"{o}tgen, and Gerhard Weikum. 2018{\natexlab{b}}.
\newblock \href {https://doi.org/10.1145/3269206.3269247} {Tequila: Temporal question answering over knowledge bases}.
\newblock In \emph{Proceedings of the 27th ACM International Conference on Information and Knowledge Management}, CIKM '18, page 1807–1810.

\bibitem[{Jia et~al.(2021)Jia, Pramanik, Saha~Roy, and Weikum}]{jia2021complex}
Zhen Jia, Soumajit Pramanik, Rishiraj Saha~Roy, and Gerhard Weikum. 2021.
\newblock Complex temporal question answering on knowledge graphs.
\newblock In \emph{Proceedings of the 30th ACM international conference on information \& knowledge management}, pages 792--802.

\bibitem[{Jiang et~al.(2023)Jiang, Sablayrolles, Mensch, Bamford, Chaplot, de~las Casas, Bressand, Lengyel, Lample, Saulnier, Lavaud, Lachaux, Stock, Scao, Lavril, Wang, Lacroix, and Sayed}]{jiang2023mistral7b}
Albert~Q. Jiang, Alexandre Sablayrolles, Arthur Mensch, Chris Bamford, Devendra~Singh Chaplot, Diego de~las Casas, Florian Bressand, Gianna Lengyel, Guillaume Lample, Lucile Saulnier, Lélio~Renard Lavaud, Marie-Anne Lachaux, Pierre Stock, Teven~Le Scao, Thibaut Lavril, Thomas Wang, Timothée Lacroix, and William~El Sayed. 2023.
\newblock \href {https://arxiv.org/abs/2310.06825} {Mistral 7b}.
\newblock \emph{Preprint}, arXiv:2310.06825.

\bibitem[{Jiang et~al.(2024)Jiang, Sablayrolles, Roux, Mensch, Savary, Bamford, Chaplot, de~las Casas, Hanna, Bressand, Lengyel, Bour, Lample, Lavaud, Saulnier, Lachaux, Stock, Subramanian, Yang, Antoniak, Scao, Gervet, Lavril, Wang, Lacroix, and Sayed}]{jiang2024mixtralexperts}
Albert~Q. Jiang, Alexandre Sablayrolles, Antoine Roux, Arthur Mensch, Blanche Savary, Chris Bamford, Devendra~Singh Chaplot, Diego de~las Casas, Emma~Bou Hanna, Florian Bressand, Gianna Lengyel, Guillaume Bour, Guillaume Lample, Lélio~Renard Lavaud, Lucile Saulnier, Marie-Anne Lachaux, Pierre Stock, Sandeep Subramanian, Sophia Yang, Szymon Antoniak, Teven~Le Scao, Théophile Gervet, Thibaut Lavril, Thomas Wang, Timothée Lacroix, and William~El Sayed. 2024.
\newblock \href {https://arxiv.org/abs/2401.04088} {Mixtral of experts}.
\newblock \emph{Preprint}, arXiv:2401.04088.

\bibitem[{Kojima et~al.(2022)Kojima, Gu, Reid, Matsuo, and Iwasawa}]{NEURIPS2022_8bb0d291}
Takeshi Kojima, Shixiang~(Shane) Gu, Machel Reid, Yutaka Matsuo, and Yusuke Iwasawa. 2022.
\newblock \href {https://proceedings.neurips.cc/paper_files/paper/2022/file/8bb0d291acd4acf06ef112099c16f326-Paper-Conference.pdf} {Large language models are zero-shot reasoners}.
\newblock In \emph{Advances in Neural Information Processing Systems}, volume~35, pages 22199--22213. Curran Associates, Inc.

\bibitem[{Naik et~al.(2019)Naik, Breitfeller, and Rose}]{naik-etal-2019-tddiscourse}
Aakanksha Naik, Luke Breitfeller, and Carolyn Rose. 2019.
\newblock \href {https://doi.org/10.18653/v1/W19-5929} {{TDD}iscourse: A dataset for discourse-level temporal ordering of events}.
\newblock In \emph{Proceedings of the 20th Annual SIGdial Meeting on Discourse and Dialogue}, pages 239--249.

\bibitem[{Ning et~al.(2020)Ning, Wu, Han, Peng, Gardner, and Roth}]{ning-etal-2020-torque}
Qiang Ning, Hao Wu, Rujun Han, Nanyun Peng, Matt Gardner, and Dan Roth. 2020.
\newblock \href {https://doi.org/10.18653/v1/2020.emnlp-main.88} {{TORQUE}: A reading comprehension dataset of temporal ordering questions}.
\newblock In \emph{Proceedings of the 2020 Conference on Empirical Methods in Natural Language Processing (EMNLP)}, pages 1158--1172.

\bibitem[{Sanampudi and Kumari(2010)}]{sanampudi2010temporal}
Suresh~Kumar Sanampudi and G~Vijaya Kumari. 2010.
\newblock Temporal reasoning in natural language processing: A survey.
\newblock \emph{International Journal of Computer Applications}, 1(4):68--72.

\bibitem[{Saxena et~al.(2021)Saxena, Chakrabarti, and Talukdar}]{saxena-etal-2021-question}
Apoorv Saxena, Soumen Chakrabarti, and Partha Talukdar. 2021.
\newblock \href {https://doi.org/10.18653/v1/2021.acl-long.520} {Question answering over temporal knowledge graphs}.
\newblock In \emph{Proceedings of the 59th Annual Meeting of the Association for Computational Linguistics and the 11th International Joint Conference on Natural Language Processing (Volume 1: Long Papers)}, pages 6663--6676.

\bibitem[{Tan et~al.(2023{\natexlab{a}})Tan, Ng, and Bing}]{Tan2023TowardsBA}
Qingyu Tan, Hwee~Tou Ng, and Lidong Bing. 2023{\natexlab{a}}.
\newblock Towards benchmarking and improving the temporal reasoning capability of large language models.
\newblock In \emph{Annual Meeting of the Association for Computational Linguistics}.

\bibitem[{Tan et~al.(2023{\natexlab{b}})Tan, Ng, and Bing}]{tan-etal-2023-towards}
Qingyu Tan, Hwee~Tou Ng, and Lidong Bing. 2023{\natexlab{b}}.
\newblock \href {https://doi.org/10.18653/v1/2023.acl-long.828} {Towards benchmarking and improving the temporal reasoning capability of large language models}.
\newblock In \emph{Proceedings of the 61st Annual Meeting of the Association for Computational Linguistics (Volume 1: Long Papers)}, pages 14820--14835.

\bibitem[{Tang and Belle(2024)}]{Tang2024LTLBenchTB}
Weizhi Tang and Vaishak Belle. 2024.
\newblock \href {https://api.semanticscholar.org/CorpusID:271051004} {Ltlbench: Towards benchmarks for evaluating temporal logic reasoning in large language models}.
\newblock \emph{ArXiv}, abs/2407.05434.

\bibitem[{Terenziani(2009)}]{terenziani2009}
Paolo Terenziani. 2009.
\newblock \href {https://doi.org/10.1007/978-0-387-39940-9_287} {\emph{Qualitative Temporal Reasoning}}, pages 2225--2229.

\bibitem[{UzZaman et~al.(2013)UzZaman, Llorens, Derczynski, Allen, Verhagen, and Pustejovsky}]{uzzaman-etal-2013-semeval}
Naushad UzZaman, Hector Llorens, Leon Derczynski, James Allen, Marc Verhagen, and James Pustejovsky. 2013.
\newblock {S}em{E}val-2013 task 1: {T}emp{E}val-3: Evaluating time expressions, events, and temporal relations.
\newblock In \emph{Second Joint Conference on Lexical and Computational Semantics (*{SEM}), Volume 2: Proceedings of the Seventh International Workshop on Semantic Evaluation ({S}em{E}val 2013)}, pages 1--9.

\bibitem[{Vashishtha et~al.(2020)Vashishtha, Poliak, Lal, Van~Durme, and White}]{vashishtha-etal-2020-temporal}
Siddharth Vashishtha, Adam Poliak, Yash~Kumar Lal, Benjamin Van~Durme, and Aaron~Steven White. 2020.
\newblock \href {https://doi.org/10.18653/v1/2020.findings-emnlp.363} {Temporal reasoning in natural language inference}.
\newblock In \emph{Findings of the Association for Computational Linguistics: EMNLP 2020}, pages 4070--4078.

\bibitem[{Vrande{\v{c}}i{\'c} and Kr{\"o}tzsch(2014)}]{vrandevcic2014wikidata}
Denny Vrande{\v{c}}i{\'c} and Markus Kr{\"o}tzsch. 2014.
\newblock Wikidata: a free collaborative knowledgebase.
\newblock \emph{Communications of the ACM}, 57(10):78--85.

\bibitem[{Wang and Zhao(2023)}]{Wang2023TRAMBT}
Yuqing Wang and Yun Zhao. 2023.
\newblock Tram: Benchmarking temporal reasoning for large language models.
\newblock \emph{ArXiv}, abs/2310.00835.

\bibitem[{Wei et~al.(2022)Wei, Wang, Schuurmans, Bosma, Xia, Chi, Le, Zhou et~al.}]{wei2022chain}
Jason Wei, Xuezhi Wang, Dale Schuurmans, Maarten Bosma, Fei Xia, Ed~Chi, Quoc~V Le, Denny Zhou, et~al. 2022.
\newblock Chain-of-thought prompting elicits reasoning in large language models.
\newblock \emph{Advances in neural information processing systems}, 35:24824--24837.

\bibitem[{Xiong et~al.(2024)Xiong, Payani, Kompella, and Fekri}]{xiong2024large}
Siheng Xiong, Ali Payani, Ramana Kompella, and Faramarz Fekri. 2024.
\newblock \href {https://arxiv.org/abs/2401.06853} {Large language models can learn temporal reasoning}.
\newblock \emph{Preprint}, arXiv:2401.06853.

\bibitem[{Yang et~al.(2023)Yang, Du, Mao, Ni, and Cambria}]{yang2023logical}
Zonglin Yang, Xinya Du, Rui Mao, Jinjie Ni, and Erik Cambria. 2023.
\newblock \href {https://arxiv.org/abs/2303.12023} {Logical reasoning over natural language as knowledge representation: A survey}.
\newblock \emph{Preprint}, arXiv:2303.12023.

\bibitem[{Yuan et~al.(2023)Yuan, Xie, Huang, and Ananiadou}]{yuan2023future}
Chenhan Yuan, Qianqian Xie, Jimin Huang, and Sophia Ananiadou. 2023.
\newblock \href {https://arxiv.org/abs/2310.01074} {Back to the future: Towards explainable temporal reasoning with large language models}.
\newblock \emph{Preprint}, arXiv:2310.01074.

\bibitem[{Zhao et~al.(2022)Zhao, Zhao, Xu, Zhang, and Jin}]{zhao-etal-2022-language}
Ruilin Zhao, Feng Zhao, Guandong Xu, Sixiao Zhang, and Hai Jin. 2022.
\newblock \href {https://doi.org/10.18653/v1/2022.findings-emnlp.147} {Can language models serve as temporal knowledge bases?}
\newblock In \emph{Findings of the Association for Computational Linguistics: EMNLP 2022}, pages 2024--2037.

\bibitem[{Zhou et~al.(2021)Zhou, Richardson, Ning, Khot, Sabharwal, and Roth}]{zhou-etal-2021-temporal}
Ben Zhou, Kyle Richardson, Qiang Ning, Tushar Khot, Ashish Sabharwal, and Dan Roth. 2021.
\newblock \href {https://doi.org/10.18653/v1/2021.naacl-main.107} {Temporal reasoning on implicit events from distant supervision}.
\newblock In \emph{Proceedings of the 2021 Conference of the North American Chapter of the Association for Computational Linguistics: Human Language Technologies}, pages 1361--1371.

\end{thebibliography}

\appendix
\section{Appendix}
\label{Appx}

\subsection{Allen Question Generation}
\label{allen-generation}
To generate the Allen questions, we take the following steps:
\begin{enumerate}[noitemsep]
\item We extract real-world event pairs from Wikidata \citep{vrandevcic2014wikidata} via SPARQL. The used Wikidata content is licensed under CC0
\footnote{\url{https://www.wikidata.org/wiki/Wikidata:Licensing}}.

\item We determine the valid Allen relation for this event pair by comparing the time intervals of these events.
\item In order to map these relations into text, we verbalize each Allen relation using the prompts as depicted in Table \ref{templates}.
\end{enumerate}

\subsubsection{Negative Samples For Allen Questions}
\label{negative-sample}

For the positive samples, we put the correct Allen relations to the \textit{Hypothesis} and set the \textit{Correctness} as \textit{True}. However, for negative samples, we choose another Allen relation (e.g. choosing the "Meets" relation instead of "Before") and set the \textit{Correctness} to \textit{False}. However, since we set the time granularity as years instead of days, generating negative samples for Allen relations presents certain challenges. For example, the "Equals" relation requires that both the start and end points of two events match exactly. When we create a negative sample for "Equals", we cannot use the "Contains" relation. This is because the second event could start later and end earlier than the first event, even if the years are the same. Since the exact days/dates of the events are not known, the information provided in the context will be ambiguous. 
To address this issue, we exclude such problematic relations from the pool of candidate relations during negative sampling.

Below we provide a list of Allen relations along with the Allen relations that are excluded from its negative sample candidates to avoid such inconclusive cases.
\begin{itemize}[noitemsep]
    \item "Equals": "Overlaps", "Contains", "During", "Overlapped-By", "Started-By", "Starts", "Finished-By", "Finishes"
    \item "Started-By": "Contains", "Overlapped-By"
    \item "Starts": "Overlaps", "During"
    \item "Finished-By": "Overlaps", "Contains"
    \item "Finishes": "During", "Overlapped-By"
    \item "Meets": "Before", "Overlaps"
    \item "Met-By": "Overlapped-By", "After"             
\end{itemize}

\subsection{Detailed Results}
\label{appx:detailed-results}
For Allen questions, we report the 0-shot, 1-shot, 3-shot, and Chain-of-Thought results in Table \ref{0-shot}, Table \ref{1-shot}, Table \ref{3-shot}, and Table \ref{CoT}. Moreover, Table \ref{abstract} includes the results for the abstract setting, where we replace the actual event names with abstract names such as "Event A" and "Event B". Table \ref{single-event} reports the results of the 0-shot, few-shot, and chain-of-thought for temporal arithmetic questions (\textit{End Timepoint}, \textit{Intermediate Timepoint} and \textit{Next Occurrence}).

\subsection{Different Prompt Variants}
\label{app-diff-prompts}
ChronoSense has different prompt variants for each question type. The templates for prompt variants can be seen in Table \ref{prompt-type-templates}. In order to show the effect of different prompts, we report the average accuracy values with standard deviation across three prompt variants in 
Table \ref{prompt-category}. Although there are cases with high standard deviation, we do not observe a relation that has consistently high values.

\subsection{Computational Budget}
We ran all experiments using HuggingFace on a single Nvidia H100 - 80GB or via the OpenAI API.
None of the experiments per model took longer than 24 hours.
The experiments via the OpenAI API caused costs of less than 100\$.

\begin{table*}
\centering
\resizebox{\textwidth}{!}{%
\begin{tabular}{|p{1.8cm}|p{3.5cm}|p{22cm}|}
\hline
\textbf{Type} & \textbf{Question} & \textbf{Template}\\
\hline Allen & Equals & Did ‘Event A' begin in the same year as ‘Event B' and end in the same year as ‘Event B'? Answer True or False. \\
\hline Allen & Before & Did ‘Event A' occur before ‘Event B' without any overlap between the two events? Answer True or False.\\
\hline Allen & After & Did ‘Event A' occur after ‘Event B' without any overlap between the two events? Answer True or False.\\
\hline Allen & Overlaps & Did ‘Event A' begin before ‘Event B' and end before ‘Event B' ended, with some overlap between the two events? Answer True or False.\\
\hline Allen &  Overlapped-By & Did ‘Event B' begin before ‘Event A' and end before ‘Event A' ended, with some overlap between the two events? Answer True or False.\\
\hline Allen & Contains & Did ‘Event A' begin before ‘Event B' began and end after ‘Event B' ended, entirely containing ‘Event B'? Answer True or False.\\
\hline Allen &  During & Did ‘Event A' begin after ‘Event B' began and end before ‘Event B' ended, being entirely contained within ‘Event B'? Answer True or False.\\
\hline Allen & Started-By & Did ‘Event B' begin in the same year as ‘Event A', but end before ‘Event A' ended? Answer True or False.\\
\hline Allen & Starts & Did ‘Event A' begin in the same year as ‘Event B', but end before ‘Event B' ended? Answer True or False.\\
\hline Allen & Finished-By & Did ‘Event B' begin after ‘Event A' began and end in the same year as ‘Event A'? Answer True or False. \\
\hline Allen & Finishes & Did ‘Event A' begin after ‘Event B' began and end in the same year as ‘Event B'? Answer True or False.\\
\hline Allen & Meets & Did ‘Event A' end in the same year as ‘Event B' began? Answer True or False.\\
\hline Allen & Met-by & Did ‘Event B' end in the same year as ‘Event A' began? Answer True or False.\\
\hline Arithmetic & End timepoint& Did ‘Event A' end in the year [start+duration]? Answer True or False. \\
\hline Arithmetic & Next occurrence& Did ‘Event A' occur again in the year [next-occurrence]? Answer True or False.\\
\hline Arithmetic & Intermediate timepoint& Was ‘Event A' happening in the year [intermediate]? Answer True or False.\\
\hline
\end{tabular}
}
\caption{\label{templates}
Templates used in ChronoSense.}
\end{table*}

\begin{table*}[t]
\resizebox{\textwidth}{!}{%
\begin{tabular}{lccccccc|c}
\toprule
&\multicolumn{1}{c}{Gemma2-9B-it} 
&\multicolumn{1}{c}{GPT-4o} &\multicolumn{1}{c}{GPT-4o-mini} & \multicolumn{1}{c}{Llama3.1-8B} & \multicolumn{1}{c}{Mistral-7B} & \multicolumn{1}{c}{Mixtral-8x7B} & \multicolumn{1}{c}{Phi-3-mini} & \multicolumn{1}{c}{Average} \\
\midrule
After & 0.28* & 0.956 & 0.882 & 0.136* & 0.918 & 0.78 & 0.566 & 0.64 \\
Before & 0.458* & 0.914 & 0.838 & 0.182 & 0.896 & 0.902 & 0.758 & 0.70 \\
Contains & 0.064* & 0.884 & 0.728 & 0.07* & 0.466 & 0.472 & 0.652 & 0.47 \\
During & 0.096* & 0.878 & 0.668 & 0.142* & 0.476 & 0.512 & 0.49 & 0.46 \\
Equals & 0.012* & 0.69 & 0.53 & 0.166* & 0.102 & 0.336 & 0.54 & 0.33 \\
Finished-By & 0.026* & 0.926 & 0.786 & 0.108* & 0.454 & 0.398 & 0.486 & 0.45 \\
Finishes & 0.03* & 0.908 & 0.602 & 0.134* & 0.416 & 0.43 & 0.492 & 0.43 \\
Meets & 0.108* & 0.91 & 0.782 & 0.146* & 0.492 & 0.74 & 0.488 & 0.52 \\
Met-By & 0.06* & 0.864 & 0.73 & 0.16* & 0.496 & 0.594 & 0.494 & 0.48 \\
Overlapped-By & 0.042* & 0.842 & 0.706 & 0.122* & 0.462 & 0.476 & 0.786 & 0.49 \\
Overlaps & 0.044* & 0.884 & 0.708 & 0.12* & 0.476 & 0.43 & 0.648 & 0.47 \\
Started-By & 0.03* & 0.896 & 0.754 & 0.12* & 0.424 & 0.578 & 0.474 & 0.46 \\
Starts & 0.042* & 0.846 & 0.748 & 0.156* & 0.45 & 0.442 & 0.492 & 0.45 \\
\midrule
Average & 0.09* & 0.87 & 0.72 & 0.13* & 0.50 & 0.54 & 0.56\\
\bottomrule
\end{tabular}
}
\caption{0-shot setting results on 13 Allen questions with explicit event names. (*) indicates a high number of unclear answers ($\geq$ 250).}
\label{0-shot}

\end{table*}
\begin{table*}
\resizebox{\textwidth}{!}{%
\begin{tabular}{lccccccc|c}
\toprule
&\multicolumn{1}{c}{Gemma2-9B-it} 
&\multicolumn{1}{c}{GPT-4o} &\multicolumn{1}{c}{GPT-4o-mini} & \multicolumn{1}{c}{Llama3.1-8B} & \multicolumn{1}{c}{Mistral-7B} & \multicolumn{1}{c}{Mixtral-8x7B} & \multicolumn{1}{c}{Phi-3-mini} &  \multicolumn{1}{|c}{Average} \\
\midrule
After & 0.922 & 0.966 & 0.898 & 0.006* & 0.846 & 0.85 & 0.882 & 0.76 \\
Before & 0.976 & 0.964 & 0.912 & 0.008* & 0.908 & 0.886 & 0.964 & 0.80 \\
Contains & 0.824 & 0.972 & 0.794 & 0.014* & 0.448 & 0.526 & 0.594 & 0.59 \\
During & 0.716 & 0.95 & 0.744 & 0.014* & 0.432 & 0.596 & 0.444 & 0.55 \\
Equals & 0.604 & 0.872 & 0.768 & 0.056* & 0.316 & 0.578 & 0.612 & 0.54 \\
Finished-By & 0.784 & 0.984 & 0.754 & 0.01* & 0.408 & 0.522 & 0.664 & 0.58 \\
Finishes & 0.464 & 0.944 & 0.518 & 0.006* & 0.358 & 0.472 & 0.416 & 0.45 \\
Meets & 0.91 & 0.932 & 0.88 & 0.022* & 0.442 & 0.494 & 0.478 & 0.59 \\
Met-By & 0.65 & 0.864 & 0.646 & 0.022* & 0.43 & 0.484 & 0.466 & 0.50 \\
Overlapped-By & 0.686 & 0.812 & 0.642 & 0.006* & 0.356 & 0.482 & 0.626 & 0.51 \\
Overlaps & 0.84 & 0.92 & 0.576 & 0.01* & 0.472 & 0.55 & 0.574 & 0.56 \\
Started-By & 0.602 & 0.99 & 0.856 & 0.008* & 0.37 & 0.46 & 0.394 & 0.52 \\
Starts & 0.884 & 0.936 & 0.856 & 0.004* & 0.418 & 0.506 & 0.56 & 0.59 \\
\midrule
Average & 0.75 & 0.93 & 0.75 & 0.01* & 0.47 & 0.56 & 0.59 \\
\bottomrule
\end{tabular}
}
\caption{1-shot setting results on Allen questions with explicit event names. (*) indicates a high number of unclear answers ($\geq$ 250).}
\label{1-shot}
\end{table*}
\begin{table*}
\resizebox{\textwidth}{!}{%
\begin{tabular}{lccccccc|c}
\toprule
&\multicolumn{1}{c}{Gemma2-9B-it} 
&\multicolumn{1}{c}{GPT-4o} &\multicolumn{1}{c}{GPT-4o-mini} & \multicolumn{1}{c}{Llama3.1-8B} & \multicolumn{1}{c}{Mistral-7B} & \multicolumn{1}{c}{Mixtral-8x7B} & \multicolumn{1}{c}{Phi-3-mini} & \multicolumn{1}{|c}{Average} \\
\midrule
After & 0.318 & 0.974 & 0.892 & 0.008* & 0.734 & 0.798 & 0.866 & 0.65 \\
Before & 0.384 & 0.966 & 0.928 & 0.014* & 0.768 & 0.944 & 0.954 & 0.70 \\
Contains & 0.222* & 0.984 & 0.832 & 0.014* & 0.45 & 0.504 & 0.512 & 0.50 \\
During & 0.218* & 0.978 & 0.774 & 0.008* & 0.46 & 0.534 & 0.42 & 0.48 \\
Equals & 0.364 & 0.948 & 0.866 & 0.046* & 0.414 & 0.52 & 0.862 & 0.57 \\
Finished-By & 0.23* & 0.98 & 0.7 & 0.004* & 0.462 & 0.512 & 0.668 & 0.50 \\
Finishes & 0.128* & 0.972 & 0.558 & 0.008* & 0.402 & 0.414 & 0.452 & 0.41 \\
Meets & 0.484 & 0.956 & 0.924 & 0.012* & 0.464 & 0.722 & 0.624 & 0.59 \\
Met-By & 0.264* & 0.89 & 0.69 & 0.018* & 0.468 & 0.558 & 0.494 & 0.48 \\
Overlapped-By & 0.194* & 0.804 & 0.63 & 0.014* & 0.454 & 0.392 & 0.732 & 0.45 \\
Overlaps & 0.152* & 0.944 & 0.518 & 0.002* & 0.474 & 0.576 & 0.74 & 0.48 \\
Started-By & 0.258* & 0.998 & 0.934 & 0.004* & 0.442 & 0.54 & 0.584 & 0.53 \\
Starts & 0.286* & 0.98 & 0.916 & 0.0* & 0.446 & 0.576 & 0.78 & 0.56 \\
\midrule
Average  & 0.26* & 0.95 & 0.78 & 0.01* & 0.49 & 0.58 & 0.66 \\
\bottomrule
\end{tabular}
}
\caption{3-shot setting results on Allen questions with explicit event names. (*) indicates a high number of unclear answers ($\geq$ 250).}
\label{3-shot}

\end{table*}
\begin{table*}
\resizebox{\textwidth}{!}{%
\begin{tabular}{lccccccc|c}
\toprule
&\multicolumn{1}{c}{Gemma2-9B-it} 
&\multicolumn{1}{c}{GPT-4o} &\multicolumn{1}{c}{GPT-4o-mini} & \multicolumn{1}{c}{Llama3.1-8B} & \multicolumn{1}{c}{Mistral-7B} & \multicolumn{1}{c}{Mixtral-8x7B} & \multicolumn{1}{c}{Phi-3-mini} & \multicolumn{1}{|c}{Average} \\
\midrule
After & 0.922 & 0.778 & 0.822 & 0.81 & 0.928 & 0.808 & 0.87 & 0.84 \\
Before & 0.956 & 0.766 & 0.848 & 0.894 & 0.918 & 0.934 & 0.932 & 0.89 \\
Contains & 0.748 & 0.666 & 0.766 & 0.682 & 0.41 & 0.676 & 0.798 & 0.67 \\
During & 0.846 & 0.684 & 0.804 & 0.732 & 0.434 & 0.636 & 0.644 & 0.68 \\
Equals & 0.452 & 0.596 & 0.672 & 0.482 & 0.248 & 0.406 & 0.576 & 0.49 \\
Finished-By & 0.69 & 0.58 & 0.446* & 0.756 & 0.444 & 0.512 & 0.794 & 0.60 \\
Finishes & 0.658 & 0.584 & 0.502 & 0.766 & 0.436 & 0.46 & 0.852 & 0.60 \\
Meets & 0.812 & 0.788 & 0.818 & 0.828 & 0.468 & 0.586 & 0.87 & 0.73 \\
Met-By & 0.69 & 0.768 & 0.814 & 0.81 & 0.532 & 0.512 & 0.85 & 0.71 \\
Overlapped-By & 0.746 & 0.542 & 0.66 & 0.66 & 0.486 & 0.468 & 0.552 & 0.58 \\
Overlaps & 0.738 & 0.504 & 0.712 & 0.726 & 0.598 & 0.488 & 0.538 & 0.61 \\
Started-By & 0.68 & 0.602 & 0.528 & 0.8 & 0.426 & 0.526 & 0.75 & 0.61 \\
Starts & 0.828 & 0.662 & 0.586 & 0.84 & 0.408 & 0.496 & 0.818 & 0.66 \\
\midrule
Average & 0.75 & 0.65 & 0.69 & 0.75 & 0.51 & 0.57 & 0.75 \\
\bottomrule
\end{tabular}
}
\caption{Chain-of-Thought setting results on Allen questions with explicit event names. (*) indicates a high number of unclear answers ($\geq$ 250).}
\label{CoT}

\end{table*}
\begin{table*}
\resizebox{\textwidth}{!}{%
\begin{tabular}{lccccccc|c}
\toprule
&\multicolumn{1}{c}{Gemma2-9B-it} 
&\multicolumn{1}{c}{GPT-4o} &\multicolumn{1}{c}{GPT-4o-mini} & \multicolumn{1}{c}{Llama3.1-8B} & \multicolumn{1}{c}{Mistral-7B} & \multicolumn{1}{c}{Mixtral-8x7B} & \multicolumn{1}{c}{Phi-3-mini} & \multicolumn{1}{c}{Average} \\
\midrule
After & 0.206* & 0.948 & 0.87 & 0.076* & 0.448* & 0.672 & 0.498 & 0.53 \\
Before & 0.498* & 0.918 & 0.82 & 0.152* & 0.458* & 0.846 & 0.742 & 0.63 \\
Contains & 0.142* & 0.848 & 0.666 & 0.038* & 0.2 & 0.38 & 0.85 & 0.44 \\
During & 0.286* & 0.806 & 0.512 & 0.062* & 0.202* & 0.33 & 0.496 & 0.38 \\
Equals & 0.0* & 0.448 & 0.382 & 0.522 & 0.22* & 0.016* & 0.692 & 0.32 \\
Finished-By & 0.024* & 0.852 & 0.736 & 0.074* & 0.182* & 0.358* & 0.5 & 0.38 \\
Finishes & 0.026* & 0.708 & 0.58 & 0.08* & 0.054* & 0.184* & 0.492 & 0.30 \\
Meets & 0.602 & 0.938 & 0.716 & 0.47 & 0.404 & 0.4 & 0.5 & 0.57 \\
Met-By & 0.062* & 0.752 & 0.682 & 0.126* & 0.306 & 0.426 & 0.494 & 0.40 \\
Overlapped-By & 0.008* & 0.552 & 0.46 & 0.034* & 0.188* & 0.216 & 0.796 & 0.32 \\
Overlaps & 0.014* & 0.77 & 0.508 & 0.032* & 0.218* & 0.292* & 0.926 & 0.39 \\
Started-By & 0.008* & 0.884 & 0.738 & 0.134* & 0.098* & 0.24 & 0.496 & 0.37 \\
Starts & 0.1* & 0.844 & 0.67 & 0.106* & 0.09* & 0.218* & 0.498 & 0.36 \\
\midrule
Average & 0.15* & 0.78 & 0.64 & 0.14* & 0.23* & 0.35 & 0.61 \\
\bottomrule
\end{tabular}
}
\caption{0-shot setting results on Allen questions with the abstract event names. (*) indicates a high number of unclear answers ($\geq$ 250).}
\label{abstract}
\end{table*}
\begin{table*}
\resizebox{\textwidth}{!}{%
\begin{tabular}{lccccccc|c}
\toprule
&\multicolumn{1}{c}{Gemma2-9B-it} 
&\multicolumn{1}{c}{GPT-4o} &\multicolumn{1}{c}{GPT-4o-mini} & \multicolumn{1}{c}{Llama3.1-8B} & \multicolumn{1}{c}{Mistral-7B} & \multicolumn{1}{c}{Mixtral-8x7B} & \multicolumn{1}{c}{Phi-3-mini} & \multicolumn{1}{c}{Average} \\
\midrule
End-Timepoint (0-shot) & 0.67 & 0.552 & 0.652 & 0.462 & 0.558 & 0.456 & 0.604 & 0.56 \\
Intermediate-Timepoint (0-shot)& 0.938 & 1.0 & 0.996 & 0.878 & 0.452* & 0.468 & 0.994 & 0.81 \\
Next-Occurence (0-shot)& 0.678 & 0.126* & 0.158* & 0.104* & 0.082* & 0.128* & 0.432 & 0.24 \\
\midrule
Average (0-shot)& 0.76 & 0.55 & 0.60 & 0.48 & 0.36* & 0.35 & 0.67\\
\midrule
End-Timepoint (1-shot)& 0.588 & 0.446* & 0.404 & 0.156 & 0.44 & 0.528 & 0.556 & 0.44 \\
Intermediate-Timepoint (1-shot) & 0.984 & 1.0 & 0.988 & 0.392 & 0.518 & 0.584 & 0.872 & 0.76 \\
Next-Occurence (1-shot) & 0.578 & 0.062* & 0.1* & 0.038 & 0.336 & 0.394 & 0.522 & 0.29 \\
\midrule
Average (1-shot)& 0.71 & 0.50* & 0.49 & 0.19 & 0.43 & 0.50 & 0.65\\
\midrule
End-Timepoint (3-shot)& 0.588 & 0.494 & 0.172* & 0.076* & 0.358 & 0.614 & 0.352 & 0.37 \\
Intermediate-Timepoint (3-shot)& 0.972 & 1.0 & 0.998 & 0.252* & 0.664 & 0.8 & 0.476 & 0.73 \\
Next-Occurence (3-shot)& 0.082* & 0.054* & 0.008* & 0.0* & 0.414 & 0.534 & 0.31 & 0.20 \\
\midrule
Average (3-shot)& 0.54 & 0.51 & 0.39* & 0.10* & 0.47 & 0.64 & 0.37\\
\midrule
End-Timepoint (CoT)& 0.992 & 0.978 & 0.978 & 0.988 & 0.798 & 0.92 & 0.996 & 0.95 \\
Intermediate-Timepoint (CoT)& 0.978 & 0.998 & 0.998 & 0.972 & 0.542 & 0.566 & 0.984 & 0.86 \\
Next-Occurence (CoT)& 0.874 & 1.0 & 1.0 & 0.82 & 0.768 & 0.788 & 0.962 & 0.88 \\
\midrule
Average (CoT)& 0.94 & 0.99 & 0.99 & 0.92 & 0.70 & 0.75 & 0.98\\
\bottomrule
\end{tabular}
}
\caption{The results on all temporal arithmetic questions in 0-, 1-, and 3-shot settings, as well as using CoT prompting. (*) indicates a high number of unclear answers ($\geq$ 250).}
\label{single-event}
\end{table*}

\begin{table*}
\centering
  \resizebox{0.8\textwidth}{!}{ 
\begin{tabular}{|p{2.5cm}|p{6cm}|p{6cm}|}
\hline
\textbf{Question} & \textbf{Prompt alternative 1} &  \textbf{Prompt alternative 2}\\
\hline Equals &  Does ‘Event A' have identical start and end years as ‘Event B'? Answer True or False.
& Are the starting and ending years of ‘Event A' and ‘Event B' the same? Answer True or False.\\

\hline Before & Is it true that ‘Event A' took place completely before ‘Event B'? Answer True or False.
& Can it be confirmed that ‘Event A' completely preceded ‘Event B'? Answer True or False.\\

\hline After & Is it true that ‘Event A' took place completely after ‘Event B'? Answer True or False.
& Can it be confirmed that ‘Event A' completely succeeded ‘Event B'? Answer True or False. \\

\hline Overlaps &  Does ‘Event A' overlap with ‘Event B' by starting before and ending during it? Answer True or False.
& Is there a period where ‘Event A' and ‘Event B' overlapped, with ‘Event A' starting and ending first? Answer True or False.\\
\hline Overlapped-By & Does ‘Event A' overlap with ‘Event B' by starting after and ending after it? Answer True or False.
&	Is there a period where ‘Event A' and ‘Event B' overlapped, with ‘Event A' starting and ending last? Answer True or False. \\

\hline Contains & Does ‘Event A' fully enclose ‘Event A', starting before and ending after ‘Event B'? Answer True or False.
&	Does the time interval of ‘Event A' contain the time interval of ‘Event B'? Answer True or False.\\
\hline During & Is ‘Event A' fully enclosed by ‘Event B', starting and ending within ‘Event B's duration? Answer True or False.
&	Can ‘Event A' be considered to occur entirely during ‘Event B', from start to finish? Answer True or False. \\

\hline Started-By &	 Does ‘Event A' have the same starting year as ‘Event B' but finish later? Answer True or False.
& Did ‘Event A' start in the same year as ‘Event B' yet end later? Answer True or False.
\\
\hline Starts &  Does ‘Event A' have the same starting year as ‘Event B' but finish earlier? Answer True or False.
& Did ‘Event A' start in the same year as ‘Event B' yet end sooner? Answer True or False.\\

\hline Finished-By  &  Does ‘Event A' start before the start of ‘Event B' and finish in the same calendar year? Answer True or False.
&	Is ‘Event A' starting earlier than ‘Event A' and concluding within the same year? Answer True or False.
\\
\hline Finishes & Does ‘Event A' start after the start of ‘Event B' and finish in the same calendar year? Answer True or False.
& Is ‘Event A' starting later than ‘Event B' and concluding within the same year? Answer True or False.
\\

\hline Meets &  Is the end of ‘Event A' coinciding with the start of ‘Event B' in the same year? Answer True or False.
&Does the end of ‘Event A' align with the beginning of ‘Event B' within the same year? Answer True or False.\\
\hline Met-by & Is the start of ‘Event A' coinciding with the end of ‘Event B' in the same year? Answer True or False.
& Does the beginning of ‘Event A' align with the end of ‘Event B' within the same year? Answer True or False. \\

\hline End timepoint & Is the conclusion of ‘Event A' marked within the year [start+duration]? Answer True or False.
&	Can it be confirmed that ‘Event A' finished in the year [start+duration]? Answer True or False.\\
\hline Next occurrence&
Is a recurrence of ‘Event A' expected in the year [next-occurrence]? Answer True or False.
& Can we anticipate another instance of ‘Event A' in the year [next-occurrence]? Answer True or False.\\
\hline Intermediate timepoint & 	During the year [intermediate], was ‘Event A' in progress? Answer True or False. & In the year [intermediate], can it be verified that ‘Event A' was active? Answer True or False. \\
\hline
\end{tabular}}
\caption{\label{prompt-type-templates}
The different prompt variants used in ChronoSense.}
\end{table*}

\begin{table*}
\resizebox{\textwidth}{!}{%
\begin{tabular}{lccccccc}
\toprule
& Gemma2-9b-it & GPT-4o & GPT-4o-mini & Llama3.1-8B & Mistral-7B & Mixtral-8x7B & Phi-3-mini \\
\midrule
After & $0.36 \pm 0.09$ & $0.95 \pm 0.00$ & $0.82 \pm 0.15$ & $0.18 \pm 0.04$ & $0.65 \pm 0.23$ & $0.67 \pm 0.18$ & $0.55 \pm 0.05$ \\
Before & $0.63 \pm 0.17$ & $0.95 \pm 0.04$ & $0.93 \pm 0.08$ & $0.47 \pm 0.25$ & $0.87 \pm 0.06$ & $0.93 \pm 0.03$ & $0.82 \pm 0.15$ \\
Contains & $0.13 \pm 0.06$ & $0.92 \pm 0.03$ & $0.60 \pm 0.16$ & $0.13 \pm 0.05$ & $0.48 \pm 0.01$ & $0.50 \pm 0.04$ & $0.67 \pm 0.21$ \\
During & $0.13 \pm 0.04$ & $0.87 \pm 0.02$ & $0.50 \pm 0.15$ & $0.17 \pm 0.02$ & $0.48 \pm 0.01$ & $0.49 \pm 0.02$ & $0.49 \pm 0.01$ \\
Equals & $0.08 \pm 0.11$ & $0.81 \pm 0.11$ & $0.41 \pm 0.13$ & $0.30 \pm 0.13$ & $0.10 \pm 0.01$ & $0.42 \pm 0.11$ & $0.51 \pm 0.03$ \\
Finished-By & $0.06 \pm 0.03$ & $0.79 \pm 0.25$ & $0.67 \pm 0.30$ & $0.16 \pm 0.05$ & $0.49 \pm 0.08$ & $0.45 \pm 0.05$ & $0.48 \pm 0.01$ \\
Finishes & $0.06 \pm 0.03$ & $0.89 \pm 0.02$ & $0.63 \pm 0.06$ & $0.16 \pm 0.02$ & $0.45 \pm 0.03$ & $0.49 \pm 0.06$ & $0.52 \pm 0.04$ \\
Meets & $0.18 \pm 0.06$ & $0.90 \pm 0.03$ & $0.75 \pm 0.04$ & $0.20 \pm 0.05$ & $0.48 \pm 0.02$ & $0.57 \pm 0.15$ & $0.48 \pm 0.00$ \\
Met-By & $0.10 \pm 0.04$ & $0.81 \pm 0.05$ & $0.66 \pm 0.06$ & $0.19 \pm 0.03$ & $0.49 \pm 0.01$ & $0.52 \pm 0.07$ & $0.47 \pm 0.03$ \\
Overlapped-By & $0.08 \pm 0.04$ & $0.68 \pm 0.19$ & $0.64 \pm 0.15$ & $0.16 \pm 0.04$ & $0.47 \pm 0.01$ & $0.46 \pm 0.01$ & $0.57 \pm 0.18$ \\
Overlaps & $0.08 \pm 0.03$ & $0.85 \pm 0.04$ & $0.78 \pm 0.09$ & $0.16 \pm 0.04$ & $0.49 \pm 0.02$ & $0.50 \pm 0.06$ & $0.59 \pm 0.11$ \\
Started-By & $0.09 \pm 0.05$ & $0.93 \pm 0.03$ & $0.74 \pm 0.09$ & $0.17 \pm 0.05$ & $0.46 \pm 0.03$ & $0.54 \pm 0.08$ & $0.64 \pm 0.14$ \\
Starts & $0.09 \pm 0.05$ & $0.86 \pm 0.01$ & $0.79 \pm 0.06$ & $0.18 \pm 0.02$ & $0.45 \pm 0.01$ & $0.47 \pm 0.02$ & $0.51 \pm 0.08$ \\
\midrule
End-Timepoint & $0.55 \pm 0.19$ & $0.58 \pm 0.10$ & $0.58 \pm 0.14$ & $0.39 \pm 0.19$ & $0.41 \pm 0.24$ & $0.50 \pm 0.06$ & $0.60 \pm 0.01$ \\
Intermediate-Timepoint & $0.64 \pm 0.26$ & $1.00 \pm 0.00$ & $0.98 \pm 0.02$ & $0.58 \pm 0.32$ & $0.29 \pm 0.22$ & $0.25 \pm 0.20$ & $0.96 \pm 0.03$ \\
Next-Occurrence & $0.56 \pm 0.11$ & $0.18 \pm 0.07$ & $0.17 \pm 0.05$ & $0.14 \pm 0.09$ & $0.16 \pm 0.07$ & $0.15 \pm 0.03$ & $0.48 \pm 0.04$ \\
\bottomrule
\end{tabular}
}
\caption{The mean accuracy and standard deviation values for three prompt variants.}
\label{prompt-category}
\end{table*}

\subsection{Qualitative Results For Failure Cases}
\label{failure-cases}
In this section, we present illustrative qualitative examples. Model outputs for selected questions are shown in Table~\ref{qualitative}. As illustrated in the table, model failures occur for several reasons. These include confusion between start and end years (Example \#1), incorrect reasoning (Examples \#3 and \#6), and calculation errors or extra calculations (Examples \#4 and \#5). In some cases, the model produces incorrect answers despite providing a correct explanation (Example \#2). Errors can also result from temporal granularity, as seen in GPT-4o’s response in an "Equals" question: \textit{"The information provided only states that both events occurred between 2016 and 2017. It does not specify the exact start and end dates for each event, so we cannot conclude that they began and ended in the same years."}

\begin{table*}
\centering
  \resizebox{\textwidth}{!}{ 
\begin{tabular}{|p{0.7cm}|p{3cm}|p{10cm}|p{10cm}|}
\hline
\textbf{\#} & \textbf{Model \& Setting} & \textbf{Question}& \textbf{Answer}\\
\hline
1& GPT-4o-mini  \newline Finishes  \newline 0-shot  \newline False
 & "The event ‘Unpacking Europe' occurred between year 2001 and year 2002. The event ‘Intercontinental Handball Cup' occurred between year 1998 and year 2002. Did ‘Unpacking Europe' end in the same year as ‘Intercontinental Handball Cup' began? Answer True or False." & "True. The event ‘Unpacking Europe' ended in 2002, and the ‘Intercontinental Handball Cup' began in 1998. Therefore, ‘Unpacking Europe' ended in the same year that ‘Intercontinental Handball Cup' began." \\
\hline

2 & Phi-3-mini  \newline Finished-By  \newline 0-shot  \newline True & "The event ‘Estonian War of Independence' occurred between year 1918 and year 1920. The event ‘Syrian National Congress' occurred between year 1919 and year 1920. Did ‘Estonian War of Independence' begin before ‘Syrian National Congress' began and end in the same year as ‘Syrian National Congress'? Answer True or False." & "False. The Estonian War of Independence began in 1918 and ended in 1920, while the Syrian National Congress took place in 1919. Although both events ended in the same year, the Estonian War of Independence began before the Syrian National Congress."\\

\hline 3 & Llama-3.1-8B  \newline Before  \newline CoT  \newline False &
"The event ‘War of the Generals' occurred between year 1911 and year 1912. The event ‘Black Power Revolution' occurred between year 1968 and year 1970. Did ‘War of the Generals' occur after ‘Black Power Revolution' without any overlap between the two events? Answer True or False." & "To determine if ‘War of the Generals' occurred after ‘Black Power Revolution' without any overlap, let's analyze the information step by step: 1. ‘Black Power Revolution' occurred between 1968 and 1970. 2. ‘War of the Generals' occurred between 1911 and 1912. 3. Since ‘Black Power Revolution' occurred between 1968 and 1970, it started after 1911 and ended before 1912. 4. Therefore, ‘War of the Generals' occurred after ‘Black Power Revolution' and there was no overlap between the two events. The answer is True."
\\

\hline
4 & Mistral-7B  \newline End-Timepoint  \newline  0-shot  \newline True& "‘Event A' started in 1948. ‘Event A' took exactly 39 years. Did ‘Event A' end in the year 1987? Answer True or False." & "False. If ‘Event A' started in 1948 and took exactly 39 years, then it would have ended in the year 1987 plus 39 years, which is 2026."\\

\hline 5 & Mistral-7B  \newline Next-occurence  \newline 1-shot  \newline False    &    "‘Event A' first occurred in year 1773. ‘Event A' occurs every 5 years. Did ‘Event A' occur again in the year 1779? Answer True or False."
 & "Answer:True. If ‘Event A' occurs every 5 years, then it would have occurred again in the year 1779 (which is 5 years after 1773)."\\
\hline 6& Mixtral-8x7B  \newline  Next-occurence  \newline CoT  \newline True & "‘Event A' first occurred in year 1555. ‘Event A' occurs every 6 years. Did ‘Event A' occur again in the year 1561? Answer True or False." &  "To determine if ‘Event A' occurred again in the year 1561, we need to see if 1561 is a multiple of 6, as we know ‘Event A' happens every 6 years. 1. Multiply 6 by any whole number to get 1561: 6 * x = 1561   To find the value of x, divide 1561 by 6:  1561 / 6 $\sim$ 260.167  Since x is a whole number (6 times 256 equals 1536), ‘Event A' does not occur in the year 1561. The answer is False."\\
\hline
\end{tabular}}

\caption{Qualitative examples for failure cases.}
\label{qualitative}

\end{table*}

\end{document}